\documentclass[conference]{IEEEtran}
\IEEEoverridecommandlockouts
\usepackage{amsmath,amssymb,amsfonts}
\usepackage{algorithmic}
\usepackage{graphicx}
\usepackage{textcomp}
\usepackage{xcolor}
\usepackage{graphicx}
\usepackage{subcaption}
\usepackage{booktabs}
\usepackage{hyperref}
\usepackage{multicol}
\usepackage{multirow}
\usepackage{comment}
\usepackage{adjustbox}
\usepackage[numbers]{natbib}
\usepackage[misc]{ifsym}

\usepackage{mwe}
\def\BibTeX{{\rm B\kern-.05em{\sc i\kern-.025em b}\kern-.08em
    T\kern-.1667em\lower.7ex\hbox{E}\kern-.125emX}}
\usepackage{tikz}
\usepackage{textcomp}
\usepackage{hyperref}
\usepackage{lipsum}

\newcommand\copyrighttext{%
  \footnotesize \textcopyright 2025 IEEE. Personal use of this material is permitted.
  Permission from IEEE must be obtained for all other uses, in any current or future
  media, including reprinting/republishing this material for advertising or promotional
  purposes, creating new collective works, for resale or redistribution to servers or
  lists, or reuse of any copyrighted component of this work in other works.
  DOI: \href{<http://tex.stackexchange.com>}{<DOI No.>}}
\newcommand\copyrightnotice{%
\begin{tikzpicture}[remember picture,overlay]
\node[anchor=south,yshift=10pt] at (current page.south) {\fbox{\parbox{\dimexpr\textwidth-\fboxsep-\fboxrule\relax}{\copyrighttext}}};
\end{tikzpicture}%
}
\begin{document}

\title{Federated Domain Generalization with \\ Latent Space Inversion}
\author{\IEEEauthorblockN{Ragja Palakkadavath\IEEEauthorrefmark{3},
Hung Le\IEEEauthorrefmark{3},
Thanh Nguyen-Tang\IEEEauthorrefmark{2},
Svetha Venkatesh\IEEEauthorrefmark{3} and
Sunil Gupta\IEEEauthorrefmark{3}}
\IEEEauthorblockA{\IEEEauthorrefmark{3}Deakin Applied Artificial Intelligence Initiative, 
Deakin University, Australia\\
Email: \{s222101652, thai.le, svetha.venkatesh, sunil.gupta\}@deakin.edu.au}
\IEEEauthorblockA{\IEEEauthorrefmark{2}Ying Wu College of Computing, New Jersey Institute of Technology, USA\\
Email: thanh.nguyen@njit.edu}}
\begin{comment}
\author{\IEEEauthorblockN{1\textsuperscript{st} Anonymous Submission}
\IEEEauthorblockA{\textit{dept. name of organization (of Aff.)}}} 

\author{\IEEEauthorblockN{1\textsuperscript{st} Ragja Palakkadavath}
%\IEEEauthorblockA{\textit{Applied Artificial Intelligence Initiative} \\
%\textit{Deakin University}\\
%s222101652@deakin.edu.au}
\and
\IEEEauthorblockN{2\textsuperscript{nd} Hung Le}
%\IEEEauthorblockA{\textit{Applied Artificial Intelligence Initiative} \\
%\textit{Deakin University}\\
%thai.le@deakin.edu.au}
\and
\IEEEauthorblockN{3\textsuperscript{rd} Thanh Nguyen-Tang}
%\IEEEauthorblockA{\textit{Ying Wu College of Computing} \\
%\textit{New Jersey Institute of Technology}\\
%thanh.nguyen@njit.edu}
\and 
\IEEEauthorblockN{4\textsuperscript{th} Svetha Venkatesh}
%\IEEEauthorblockA{\textit{Applied Artificial Intelligence Initiative} \\
%\textit{Deakin University}\\
%svetha.venkatesh@deakin.edu.au}
\and 
\IEEEauthorblockN{5\textsuperscript{th} Sunil Gupta}
\IEEEauthorblockA{\textit{Applied Artificial Intelligence Initiative} \\
\textit{Deakin University}\\
sunil.gupta@deakin.edu.au}}
\end{comment}
\maketitle
\copyrightnotice
\begin{abstract}
Federated domain generalization (FedDG) addresses distribution shifts among clients in a federated learning framework. FedDG methods aggregate the parameters of locally trained client models to form a global model that generalizes to unseen clients while preserving data privacy. While improving the generalization capability of the global model, many existing approaches in FedDG jeopardize privacy by sharing statistics of client data between themselves. Our solution addresses this problem by contributing new ways to perform local client training and model aggregation. To improve local client training, we enforce (domain) invariance across local models with the help of a novel technique, \textbf{latent space inversion}, which enables better client privacy. When clients are not \emph{i.i.d}, aggregating their local models may discard certain local adaptations. To overcome this, we propose an \textbf{important weight} aggregation strategy to prioritize parameters that significantly influence predictions of local models during aggregation. Our extensive experiments show that our approach achieves superior results over state-of-the-art methods with less communication overhead. Our code is available \href{https://github.com/ragjapk/Latent-Space-Inversion}{here}.
\end{abstract}


\section{Introduction}
\label{sec:intro}
In many real-world applications, when developing predictive models, restrictions in data sharing make it challenging to train a single model in a distributed setup. For example, when developing a model to analyze healthcare data from multiple sources, healthcare providers face restrictions on sharing patient information among themselves or storing it in a central repository because it contains sensitive information. Federated learning (FL)~\citep{mcmahan2017communication} allows multiple distributed clients to collaboratively train a global model by sharing the parameters of their local models with a central server while keeping their local data private. The central server aggregates the model parameters to compute the global model. However, differences in data distribution across local client data can degrade the global model performance in standard federated learning methods. 

Meanwhile, domain generalization (DG)~\citep{NIPS2011_b571ecea} techniques leverage data from source domains with potentially different distributions to extract a domain-agnostic model that generalizes predictive performance from these domains to an unseen domain. DG enables the model to generalize across source and unseen domains, but it assumes a centralized setting where data from all source domains is available in one place. 

\begin{figure}[!t]
\centering
\includegraphics[width=\linewidth]{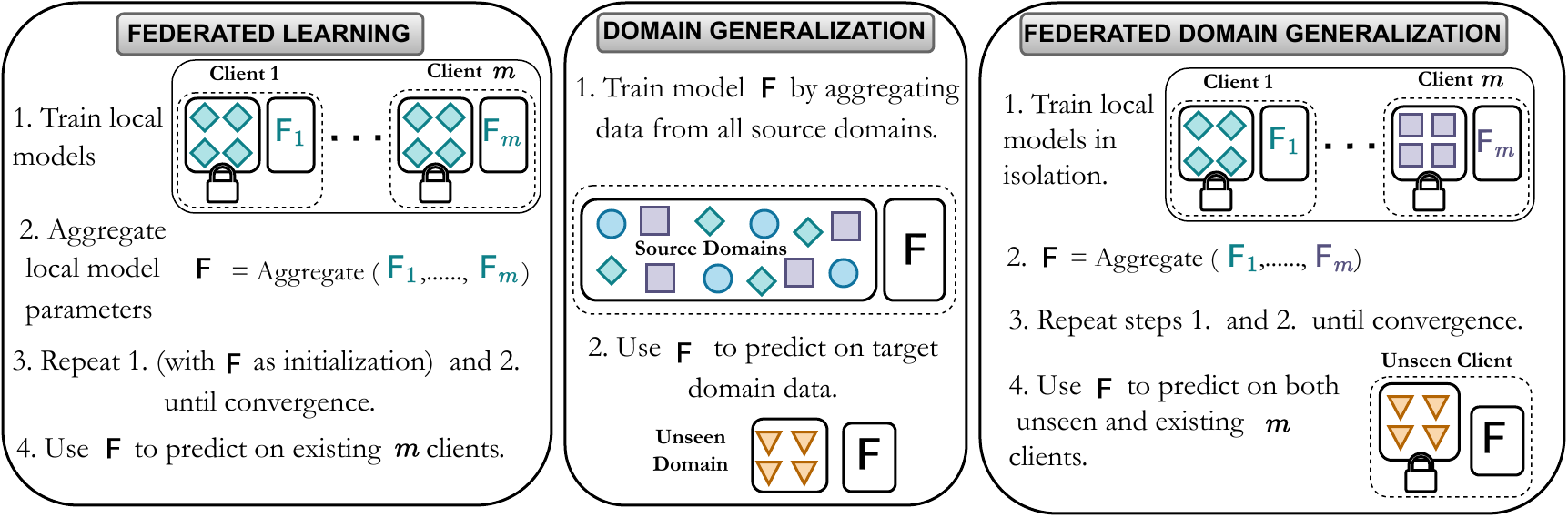}
\caption{Illustration of a model $\mathsf{F}$ trained in federated learning, domain generalization, and federated domain generalization.}
\label{fig:graph}
\end{figure}

A key challenge in the deployment of DG algorithms in real-world settings arises when data providers (such as hospitals or financial institutions) do not share raw data due to privacy, legal, or infrastructural constraints. This precludes training a single model using centralized data, which is often required to establish domain invariance between data from different domains. Federated domain generalization (FedDG) bridges this gap by incorporating generalization techniques into the federated learning framework. Each client in FedDG may have a potentially different data distribution from the others. FedDG techniques are expected to capture domain invariance between clients and generalize the global model to an unseen client without violating client data privacy. In this work, we consider that each federated client comes from a different domain distribution. Therefore, we use the terms \textit{client} and \textit{domain} interchangeably. Fig.~\ref{fig:graph} illustrates the training of a model $\mathsf{F}$ across federated learning, domain generalization, and federated domain generalization. 

Recently, many approaches~\cite{le2024efficiently,chen2023federated,park2023stablefdg} have tried to develop efficient solutions for FedDG. Of these, CCST~\cite{chen2023federated} and StableFDG~\cite{park2023stablefdg} rely on sharing data statistics (such as style information) between clients, which may pose privacy risks as per~\cite{shao2024surveysharefederatedlearning}. gPerXAN~\cite{le2024efficiently} does not share data directly, but it relies on batch normalization statistics from all layers of local models. Model inversion~\cite{yin2020dreaming} can utilize normalization statistics from all layers to recover original training data, compromising client privacy. Even though the synthetic images are artificially generated, they may closely resemble original images and inadvertently reveal sensitive information. 

Domain invariant representation learning~\cite{NEURIPS2021_2a271795,arjovsky2019invariant} is an effective paradigm in DG, allowing remarkable generalization capability to unseen domains. The model is trained to learn latent representations of data that remain consistent across different domains while disregarding domain-specific variations for generalization. These representations are then used for model prediction. This concept is effective in centralized domain generalization, where data from multiple domains can be merged into a single training set to learn domain invariant representations. It cannot be directly adopted in a federated learning framework where client data cannot be exchanged or shared. However, local models can be inverted to synthesize the data originally used to train them~\citep{yin2020dreaming}. Synthesized data from each client can be propagated to the server and used to learn domain invariance. Using synthesized data to learn domain invariance presents two key challenges: (i) Privacy concerns – even though artificially generated, synthetic images may closely resemble original images and inadvertently reveal sensitive information. %(ii) Data quality and diversity – The effectiveness of synthesized images in learning domain-invariant representations depends on their diversity. If the generated data lacks variability, it may fail to capture the original client data distribution. 
(ii) Computational overhead – generating high-quality synthetic images can be resource intensive, accumulate high costs, and slow the training process. To overcome (i) and (ii), we propose \emph{latent space inversion}, using which we synthesize meaningful latent representations of images instead of full images, thus reducing the risk of privacy leakage. 
Another challenge of FedDG is that aggregating local models trained on non-\emph{i.i.d} data may ignore client-specific information learned during local training. To address this, we propose an aggregation strategy that assigns greater weight to parameters that are most critical to the predictions of the local model. 
Our contributions are as follows.
\begin{itemize}
\item We propose \emph{latent space inversion} in which part of a classification model is inverted to synthesize latent representations of the data originally used to train the model.
\item We demonstrate that representations synthesized through latent space inversion can facilitate learning domain invariance, enhancing model generalization without sharing actual client data.
\item We propose a new aggregation scheme called \emph{important weight} aggregation to improve generalization and personalization of the global model. 
\item We evaluate our method on three benchmark datasets in FedDG and show that our method achieves superior generalization accuracy with less communication overhead compared to the state-of-the-art methods in FedDG.
\end{itemize}
\section{Related Works}
\paragraph{\underline{\smash{Federated Learning}}}
Federated learning~\citep{mcmahan2017communication} (FL) is a machine learning paradigm where multiple distributed clients interact with a central server. Each client has a local training data set and a local model; however, they perform the same tasks (for example, classification). Local data either contains sensitive information that cannot be shared with the central server or is too large to be accommodated at the central server for training a global model. FedAvg~\citep{mcmahan2017communication} proposes to let clients perform their local computations first, then the server averages the model weights. This procedure is repeated for many iterations until convergence. In some recent works~\cite{10.1145/3625558,yu2022heterogeneous,Huang_2022_CVPR}, clients are considered heterogeneous, i.e., client data are not identically distributed. However, generalizing the global model to a client having a completely new data distribution is not addressed in standard FL algorithms. 
\paragraph{\underline{\smash{Domain Adaptation and Domain Generalization}}}
Domain adaptation~\citep{long2015learning,ganin2016domain,Csurka2017} techniques assume that unlabeled samples or in some cases a few labeled samples of the unseen distribution are available for training. It is a framework that addresses distribution shifts in which we have some prior information on the unseen distribution. Domain generalization (DG) techniques~\citep{NIPS2011_b571ecea} assume that data from unseen domains is not available during training. Instead, labeled data from distinct sources with some common characteristics is available. The approach aims to use these sources to learn a model that generalizes to a target domain that is not present among the sources. DG techniques can be broadly categorized into domain invariant representation learning~\citep{NEURIPS2021_2a271795,robey2021model,zhou2020deep,arjovsky2019invariant}, data augmentation~\citep{https://doi.org/10.48550/arxiv.1805.12018,https://doi.org/10.48550/arxiv.2008.12205,Gong2019DLOWDF}, and training strategies, including meta-learning~\citep{NEURIPS2018_647bba34} and model fusion~\citep{https://doi.org/10.48550/arxiv.1806.05810}. Most studies in DG assume that data from multiple domains are concurrently accessible to the model.

\paragraph{\underline{\smash{Federated Domain Generalization}}}
Federated domain generalization~\citep{li2024federateddomaingeneralizationsurvey} (FedDG)  is an emerging research area that combines federated learning with domain generalization. Clients in the federated learning framework can be seen as domains with different data distributions. As in federated learning, local models are trained on the local data and aggregated on the server. However, the global model must also generalize to a new client whose data distribution differs from the original clients. Recent works in this direction are as follows. FedSR~\citep{NEURIPS2022_fd946a6c} learns a simple representation of the data by minimizing the squared Euclidean ($\ell_2$) norm of the data representations of each client and reducing the mutual entropy between the data and their corresponding representations. MCGDM~\cite{wei2024multisource} employs intra-domain and inter-domain gradient matching to enforce domain invariance. FedDG-GA~\citep{Zhang_2023_CVPR} considers the aggregation weights of the local models as learnable parameters and uses them to reduce the generalization gap between the global model and the local model. In contrast to these works, we take an orthogonal approach that matches \emph{latent representations of the input} across clients to capture domain invariance and common knowledge of classes. Our method overcomes the unavailability of centralized input features by synthesizing latent representations from local models.

In StableFDG~\citep{park2023stablefdg}, clients share input style statistics among themselves to improve diversity, while in CCST~\citep{chen2023federated}, to enforce domain invariance. ELCFS~\citep{liu2021feddg} transforms the input image features into the Fourier space and divides them into phase and amplitude components. Then, amplitude information is exchanged across clients through the central server so that each client learns a multi-distribution. These works share data statistics (for example, style or amplitude) that correspond directly to input data, raising concerns about potential privacy leakage. gPerXAN~\citep{le2024efficiently} combines instance and batch normalization layers using an explicit differential mixture and introduces a regularization loss to improve domain invariance. However, gPerXAN requires sharing batch normalization statistics with the server. Exposing batch normalization statistics from every layer of the model makes it vulnerable to model inversion attacks~\cite{yin2020dreaming}. Our method only uses latent representations of the data, synthesized using batch normalization statistics from only a \emph{single} layer of the model, ensuring stronger privacy protection. 

\paragraph{\underline{\smash{Model Inversion}}}
Model inversion focuses on synthesizing images from the classification network used to train it. The authors of~\citep{10.1145/2810103.2813677} proposed a model inversion attack to obtain class images from a network by optimizing the input via gradient descent. %Following this, several other works~\citep{yang2019adversarialneuralnetworkinversion,} have focused on extracting images using the classification model. 
However, this method was demonstrated on shallow networks and required additional information (input features). The authors of~\citep{mahendran2016visualizing,mahendran2015understanding} explore inversion, activation maximization, and caricaturization to synthesize natural pre-images from a trained network. These methods still rely on auxiliary dataset information or additional pre-trained networks. One of the pioneering methods that perform inversion of a large model (for example, Resnet50~\citep{he2015deepresiduallearningimage}) is DeepInversion~\citep{yin2020dreaming}. DeepInversion optimizes random noise to match the batch normalization statistics present in each layer of the network. The optimization transforms the random noise into images originally used to train the network. Subsequent works~\citep{ijcai2021p327,yin2021see} have advanced the state of the art in synthesizing images from classifiers. Due to privacy concerns, we do not directly use DeepInversion to create synthetic images. Instead of images, we synthesize intermediate representations from the classifier that captures label information, enabling domain invariant representation learning and facilitating generalization.

\section{Methodology}
\paragraph{\underline{\smash{Problem Setting}}}
A central server coordinates a set $\mathcal{D}$ of $m$ local clients, where each client is denoted by $d \in \mathcal{D}$. Each client $d$ consists of a labeled local dataset $\mathcal{S}_d$ following a distribution $\mathcal{P}_d$, containing a collection of the form $(\mathbf{x},y)$, where $\mathbf{x} \in \mathcal{X}$ is an image and $y \in \mathcal{Y}$ is its corresponding class label, and a local prediction model. Each local model is a composite function ($\circ$) of an encoder $g_d$ and a classifier $h_d$, such that $\mathsf{F}_d=g_d \circ h_d$. %The local model is a composite function of an encoder $E$ followed by a classifier $C$. 
Due to privacy regulations, clients do not share $\mathcal{S}$ among themselves or with the central server. Instead, a global model $\mathsf{F}=(g \circ h)$ is developed as follows: 
\begin{enumerate}
    \item Each client $d$ trains their local model $(g_d,h_d)$ and sends its parameters to the central server.
    \item The server aggregates these parameters to obtain the global model $(g,h)$.
    \item The new global model parameters are shared with the clients.
    \item Each client initializes its model with the global model parameters and then retrains it.
\end{enumerate}
This process is repeated until the global model converges. An example of this is the popular algorithm FedAvg~\citep{mcmahan2017communication}. However, FedAvg is not very effective when clients are susceptible to covariate shift, i.e., image features may shift between clients. %The global model must generalize across all $k$ clients in the framework and also adapt to any new clients with different data distributions. 
\paragraph{\underline{\smash{Motivation for Our Method}}}
The following challenges arise when developing a generalizable federated global model: 
\begin{itemize}
    \item Due to distribution shifts between clients, directly averaging local models would lead to divergent model updates~\citep{zhao2018federated}.
    \item The absence of centralized training data prevents us from enforcing standard DG techniques, such as domain invariance or alignment, greatly limiting the ability of the global model to generalize to unknown clients.
    \item Aggregating local model parameters could lead to a loss of information specific to the client distributions in the global model, especially if the clients had distribution shifts or the aggregation mechanism could not effectively preserve relevant local information.
\end{itemize}
We address the first two challenges by minimizing the divergence between local models by enforcing domain invariance between them. To achieve that, we propose to synthesize meaningful latent representations of the input data. We argue that synthetic data in the form of latent representations can be pooled together to learn domain invariance between clients without compromising client privacy. To address the last challenge, we propose a new parameter aggregation strategy that replaces standard aggregation. Each parameter in the local model is weighted in proportion to its contribution to the model's prediction.
\begin{figure*}[!t]
    \centering
    \includegraphics[width=0.7\linewidth]{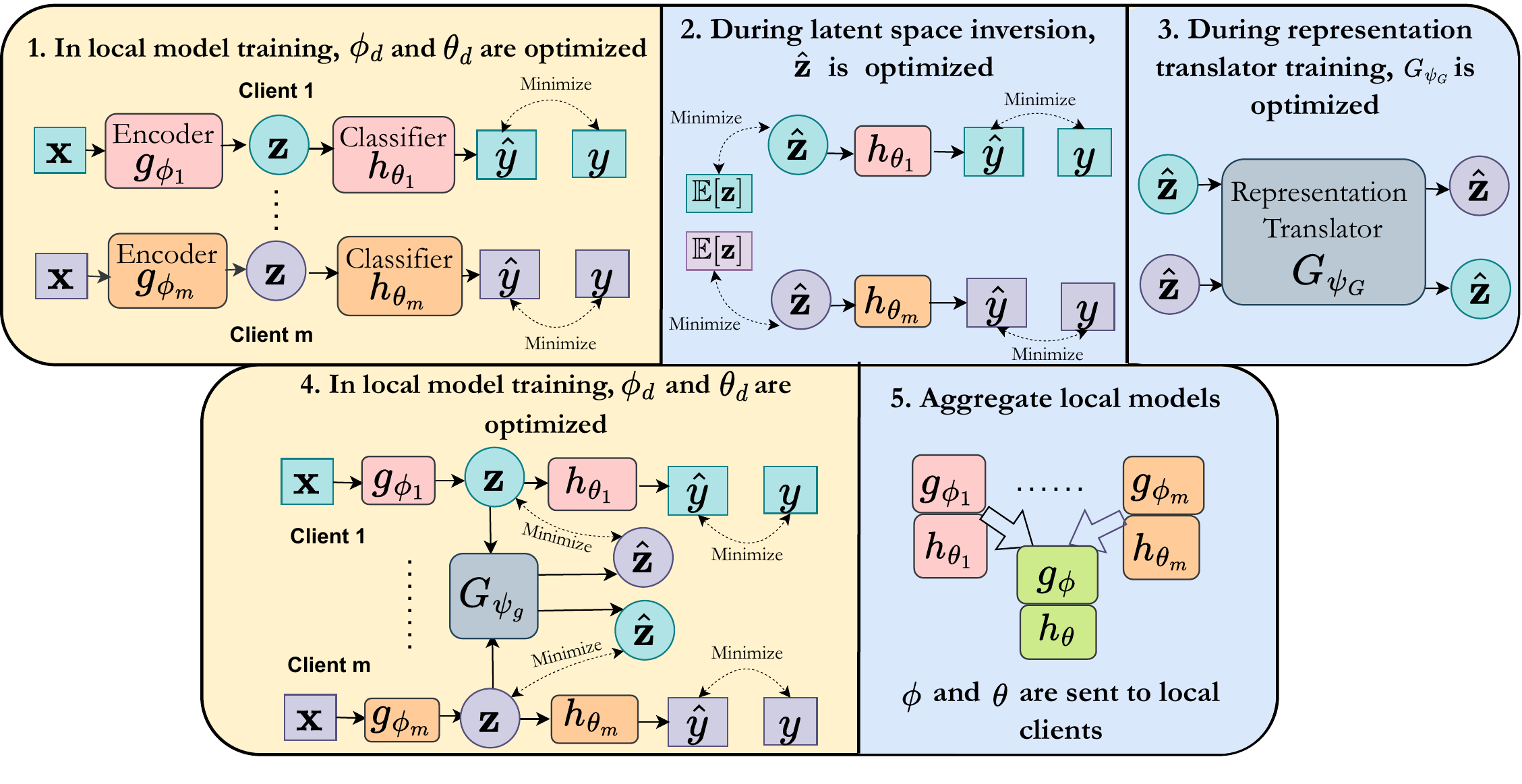}
    \caption{The five-stage framework: yellow indicates client-side stages and blue indicates server-side stages. In stage $1$, each client $d \in \mathcal{D}$ trains its local model (encoder $g_{\phi_d}$ and classifier $h_{\theta_d}$) in isolation. In stage $2$, we use $h_{\theta_d}$ to optimize noise $\mathbf{\hat{z}}$ to match original latent representations through latent space inversion. In stage $3$, we use the synthesized latent representations to train a generative model ($G$) that translates the representation from one client distribution to another while preserving its class. In stage $4$, each client trains their model with an additional loss, minimizing the divergence between the client representations and their client-translated versions. In stage $5$, local model parameters are aggregated to compute the global model. The global model then replaces the local model parameters. Stages $4$ and $5$ are repeated until the global model converges.}
    \label{fig:31}
\end{figure*}
\subsection{Design}
Our framework has $5$ stages, as depicted in Fig~\ref{fig:31}. In stage $1$, each local client trains its model and transfers only the final layer of each local model (classifier head) to the central server. In stage $2$, we employ our approach, latent space inversion, to synthesize latent representations using the classifier. In stage $3$, we use the synthesized representations to train a representation translator, which can translate representations from one client distribution to another. The server then transfers the representation translator to each local client. In stage $4$, we train each local client with an additional divergence loss that minimizes the distance between the original latent representations and their client-translated versions generated by the representation translator (enforcing invariance between latent representations of different clients). In stage $5$, the local model parameters are aggregated using a strategy that assigns a greater weight to the parameters that contribute more to the model predictions. Stages $4$ and $5$ are repeated until the global model converges. We describe each stage in detail below.          
%Encoder $g$ takes in input image $\mathbf{x}$ and outputs a latent representation $\mathbf{z}$. Classifier $h$ takes in the representation $\mathbf{z}$ as its input and outputs the final prediction $y$.  

\subsubsection{\textbf{Stage 1 - Training Local Models}}
%Training data at each client $m \in M$ is of the form $\mathcal{S}_m = \{\mathbf{x},y\}_{i=1}^{N_m} \sim \mathcal{P}_m$ where $\mathbf{x} \in \mathcal{X}$ is an image and $y \in \mathcal{Y}$ is its corresponding class label. Each client prediction model $f_m$ is a composite function of an encoder $g_m$ followed by a classifier $h_m$. Each local model is trained in isolation. 
At each client $d$, an encoder $g_{\phi_d}: \mathcal{X} \rightarrow\mathcal{Z}$, parameterized by $\phi$ takes an image $\mathbf{x}\in\mathcal{S}_d$ as input and outputs a latent representation $\mathbf{z}$. The representation lies in the latent space $\mathcal{Z}$ with a dimension of $p$. A classifier $h_{\theta_d}:\mathcal{Z} \rightarrow \mathcal{Y}$, parameterized by $\theta$ inputs the representation $\mathbf{z}$ and outputs the prediction $\hat{y}$. Thus,   
\begin{align}\label{eq:31}
\mathbf{z} &= g_{\phi_d}(\mathbf{x}) \\
\hat{y} &= h_{\theta_d}(\mathbf{z})
\end{align}
We minimize the prediction error at each client through the classification loss by optimizing parameters $\phi$ and $\theta$:
\begin{align}\label{eq:32}
\mathcal{L}_{\text{cls}}(\theta_d,\phi_d)&= \mathbb{E}_{(\mathbf{x},y)\sim \mathcal{P}_d}l(\hat{y},y)
\end{align}
where $l: \mathcal{Y} \times \mathcal{Y} \rightarrow \mathbb{R}$, (we use cross entropy loss as $l$).
\subsubsection{\textbf{Stage 2 - Latent Space Inversion}}
At this stage, we require local models to share only a partial set of their parameters with the server. This partial set contains only classifier parameters, $h_{\theta_d}$, that is, parameters of the final layer of the model. $h_{\theta_d}$ consists of a single fully connected layer preceded by a batch normalization layer. Unlike past works, this single layer of the model alone is insufficient to reconstruct images. Instead, our approach inverts it to synthesize latent representations $\mathbf{\hat{z}}$ corresponding to each client. Ideally, $\mathbf{\hat{z}}$ should be informative of any class label, closely resemble original representations $\mathbf{z}$, not reveal sensitive information, and be diverse.

We initialize $\mathbf{\hat{z}}$ as random noise, $\mathbf{\hat{z}} \sim \mathcal{N}(\mathbf{0},\mathbf{1})$ and optimize it as follows. To incorporate class information in $\mathbf{\hat{z}}$, we optimize $\mathbf{\hat{z}}$ to encourage the classifier to assign it to a set of class labels sampled from $\mathcal{S}_d$. We apply this procedure to the classifier of each client $d$.   
\begin{align}\label{eq:33}
\mathcal{L}_{\text{clsz}}(\mathbf{\hat{z}})&= \mathbb{E}_{(\mathbf{\hat{z}}\sim \mathcal{N}(\mathbf{0},\mathbf{1}), y \sim \mathcal{P}_d)}l(h_{\theta_d}(\mathbf{\hat{z}}),y)
\end{align}
To increase the similarity between original and synthesized representations, we optimize the feature map statistics of $\mathbf{\hat{z}}$ to lie close to the original $\mathbf{z}$. We assume that the original feature statistics follow the normal distribution and represent them as functions of $\mu(.)$ and $\sigma^2(.)$.  
\begin{align}\label{eq:34}
\mathcal{L}_{\text{bn}}(\mathbf{\hat{z}}) = \mathrm{Div}(\mu(\mathbf{\hat{z}}),\mathbb{E}(\mu(\mathbf{z}))) + \mathrm{Div}(\sigma^2(\mathbf{\hat{z}}), \mathbb{E}(\sigma^2(\mathbf{z})))
\end{align}
Here, $\mu(\mathbf{\hat{z}})$ and $\sigma^2(\mathbf{\hat{z}})$ are the batch-wise mean and variance estimates of $\mathbf{\hat{z}}$ and $\mathrm{Div}(\cdot)$ is a distance metric that needs to be minimized. Due to privacy concerns, feature maps of $\mathbf{z}$: $\mathbb{E}(\mu(\mathbf{z}))$ and $\mathbb{E}(\sigma^2(\mathbf{z}))$ are not available at the server. So, we replace them with the batch normalization statistics from the only classifier layer. The knowledge of the batch normalization statistics from a single layer is insufficient for model inversion~\citep{choi2021qimera}, which prevents the reconstruction of the original images and thus preserves privacy.   

We apply squared Euclidean norm ($\ell_2$ norm) regularization to synthesized representations to encourage them to be more uniformly distributed and to avoid restricting the data to a few modes. %Fig.~\ref{fig:reps} demonstrates that the regularization promotes diverse representations by discouraging the model from relying too much on a few dominant modes.
\begin{align}\label{eq:35}
\mathcal{L}_{\text{norm}}(\mathbf{\hat{z}}) = \parallel\mathbf{\hat{z}}\parallel_2^2 
\end{align}
The combined optimization objective is as follows:
\begin{align}\label{eq:36}
 \mathcal{L}_{\text{synth}}(\mathbf{\hat{z}}) = \mathcal{L}_{\text{clsz}}(\mathbf{\hat{z}}) + \lambda_{\text{bn}}*\mathcal{L}_{\text{bn}}(\mathbf{\hat{z}}) + \lambda_{\text{norm}} \:* \mathcal{L}_{\text{norm}}(\mathbf{\hat{z}}) 
\end{align}
$\lambda_{\text{bn}}$ and $\lambda_{\text{norm}}$ are coefficients of the respective losses and considered as hyperparameters.
\subsubsection{\textbf{Stage 3 - Training the Representation Translator}}
Using domain density translation to learn domain invariant representations is a popular approach in (centralized) domain generalization~\citep{NEURIPS2021_2a271795}. The authors propose enforcing representations to be invariant under all transformation functions among domains, aiming to capture domain-invariant information. However, the domain density translator \emph{operates on images}. Since we do not have access to training images of local clients due to privacy constraints, we develop a domain representation translator that \emph{translates synthetic representations ($\mathbf{\hat{z}}$) from one client distribution to another}. Let $d$ and $d'$ denote an arbitrary pair of local client indices. We learn a mapping function $G_{\psi_G}: \mathcal{Z} \times \mathbb{N} \times \mathbb{N} \rightarrow  \mathcal{Z}$, parameterized by $\psi_G$ that translates $\mathbf{\hat{z}} \in d$ to $\mathbf{\hat{z}}' \in d'$ while preserving its class $y$,
\begin{align}\label{eq:37}
\mathbf{\hat{z}}' = G_{\psi_G}(\mathbf{\hat{z}},d,d'\mid y). 
\end{align}
%, we learn a generative model to transfer recovered intermediate data representations $\mathbf{\hat{z}}$ from one domain to another. We refer to the generator of this generative model as a domain-density translator.The discriminator generates probability distributions for (i) determining whether the input is real or fake and (ii) identifying the client source.
We modeled this mapping function using a StarGAN~\citep{choi2018stargan} following DIRT~\cite{NEURIPS2021_2a271795}, in the representation space instead of the pixel space. It has the following components: a generator $G_{\psi_G}$, which accepts a latent representation $\mathbf{\hat{z}}$ of a client $d$ as input and translates it to another client, $d'$, a discriminator $D$ with two output heads: (i) $D_{\psi_{\text{src}}}$, parameterized by $\psi_{\text{src}}$, predicts if the input is \emph{real} or \emph{fake} and (ii) $D_{\psi_{\text{cls}}}$, parameterized by $\psi_{\text{cls}}$, predicts the client index of the input. 

To make the representations generated by $G_{\psi_G}$ indistinguishable from $\mathbf{\hat{z}}$, we adopt the following adversarial objective. This objective is minimized by $G_{\psi_G}$ and maximized by $D_{\psi_{\text{src}}}$ to compete against each other:
\begin{multline}\label{eq:38}
\mathcal{L}_{\text{adv}}(\psi_G,\psi_{\text{src}})=\mathbb{E}_{\mathbf{\hat{z}}}\:\text{log}[D_{\psi_{\text{src}}}(\mathbf{z})]+\\\text{log}[(1-D_{\psi_{\text{src}}}(G_{\psi_G}(\mathbf{\hat{z}},d,d')))].    
\end{multline}
Next, we apply two client classification losses. The first loss optimizes $D_{\psi_{\text{cls}}}$ to predict the correct client index of $\mathbf{\hat{z}}$, $d$:
\begin{align}\label{eq:39}
\mathcal{L}_{\text{clsd}}(\psi_{\text{cls}})=\mathbb{E}_{(\mathbf{\hat{z}},d)}-\text{log}[D_{\psi_{\text{cls}}}(d \mid \mathbf{\hat{z}})].
\end{align}
The second loss optimizes $G$ to generate representations from $d'$ that $D_{\psi_{\text{cls}}}$ correctly assigns to index $d'$: 
\begin{align}\label{eq:310}
\mathcal{L}_{\text{clsg}}(\psi_G)=\mathbb{E}_{(\mathbf{\hat{z}},d')}-\text{log}[D_{\psi_{\text{cls}}}(d' \mid G_{\psi_G}(\mathbf{\hat{z}},d,d'))].
\end{align}
Finally, a reconstruction loss is applied to preserve the content of generated representations during translation, modifying only the client index.
\begin{align}\label{eq:311}
\mathcal{L}_{\text{rec}}(\psi_G)=\mathbb{E}_{(\mathbf{\hat{z}},d)}\parallel \mathbf{\hat{z}}-G_{\psi_G}(G_{\psi_G}(\mathbf{\hat{z}},d,d'),d',d) \parallel_1,   
\end{align}
where $d' \sim \mathcal{D}$ and $\lVert \cdot\rVert_1$ is the $\ell_1$ regularization. We train the generator and the discriminator of the StarGAN by combining the losses defined above as follows:
\begin{align}\label{eq:312}
\mathcal{L}_{\text{gen}}(\psi_G)=\mathcal{L}_{\text{adv}}(\psi_G)+\lambda_{\text{clsg}}*\mathcal{L}_{\text{clsg}}(\psi_G) + \lambda_{\text{rec}}*\mathcal{L}_{\text{rec}}(\psi_G)
\end{align}
\begin{align}\label{eq:313}
\mathcal{L}_{\text{disc}}(\psi_{\text{src}},\psi_{\text{cls}})=-\mathcal{L}_{\text{adv}}(\psi_{\text{src}}) + \lambda_{\text{clsd}}*\mathcal{L}_{\text{clsd}}(\psi_{\text{cls}}).  
\end{align}
$\lambda_{\text{clsg}}$, $\lambda_{\text{rec}}$, and $\lambda_{\text{clsd}}$ are coefficients of the respective losses. We keep their values the same as that of DIRT~\cite{NEURIPS2021_2a271795}.
In practice, we enforce $G$ to transform the data distribution within the class $y$ by sampling each minibatch with data from the same class $y$, so that the discriminator will distinguish the transformed representations from the input representations from class $y$. %However, we found that this constraint can be relaxed in practice, and the generator almost always transforms the representation within the original class y.
\subsubsection{\textbf{Stage 4 - Training Local Models using the Representation Translator}}
The representation translator is transferred to every local client. Then, clients encode client-invariant behavior in their models by minimizing the divergence between the synthesized representations and translated representations as given below. 
\begin{align}\label{eq:314}
\mathcal{L}_{\text{di}}(\phi_d) = \mathbb{E}_{d \in \mathcal{D}}\mathrm{Div}(g_{\phi_d}(\mathbf{x}),G_{\psi_G}(g_{\phi_d}(\mathbf{x}),d,d')),
\end{align}
where $d'\sim \mathcal{D}$ and $\mathrm{Div}(\cdot)$ is a distance measure. We compute the final loss at each client as follows:
\begin{align}\label{eq:315}
\mathcal{L}_{\text{final}}(\theta_d,\phi_d) =\mathcal{L}_{\text{cls}}(\theta_d,\phi_d)+ \lambda_{\text{di}} * \mathcal{L}_{\text{di}}(\phi_d).
\end{align}
Here, $\lambda_{\text{di}}$ is a hyperparameter and the coefficient corresponding to the domain invariance loss. 
\subsubsection{\textbf{Stage 5 - Aggregation with Important Weights}} We adapt the technique to protect network parameters during sequential learning of tasks in continual learning to protect parameters of local modes during aggregation in FedDG. During local model training, we also compute the impact of each parameter on the model output. Given the $d^\text{th}$ local model, we denote $\omega_{d}^{j}$ to measure the impact of its $j^{\text{th}}$ parameter on its output. Inspired from~\citep{aljundi2018memoryawaresynapseslearning}, we compute $\omega_{d}^{j}$ as the rate of change in the magnitude of the model's output for a small perturbation in its parameters.
\begin{equation}\label{eq:316}
%\begin{split}
\omega^{j}_{\phi_d} =\mathbb{E}_{\mathbf{x}\sim \mathcal{P}_d}\frac{\partial\lVert g_{\phi_d}(\mathbf{x})\rVert_2}{\partial_{\phi_d^{j}}}, \omega^{j}_{\theta_d}=\mathbb{E}_{\mathbf{x}\sim \mathcal{P}_d}\frac{\partial\lVert(h_{\theta_d}(g_{\phi_d}(\mathbf{x}))\rVert_2}{\partial_{\theta_d^{j}}}  
%\end{split}
\end{equation}
Then, every $\omega_d$ is normalized across the $d$ clients, to sum to $1$. Then, the parameters are weighted by the normalized $\omega_{d}$ before being sent to the server to be aggregated.
\begin{align}\label{eq:317}
\phi= \frac{1}{m}\sum_{d=1}^{m} \omega_{\phi_d}  * \phi_d, \: \:
\theta= \frac{1}{m}\sum_{d=1}^{m} \omega_{\theta_d}  * \theta_d
\end{align}
\begin{figure*}[t]
\centering
\captionsetup[subfigure]{justification=centering}
\begin{minipage}{0.24\linewidth}
\includegraphics[width=\linewidth]{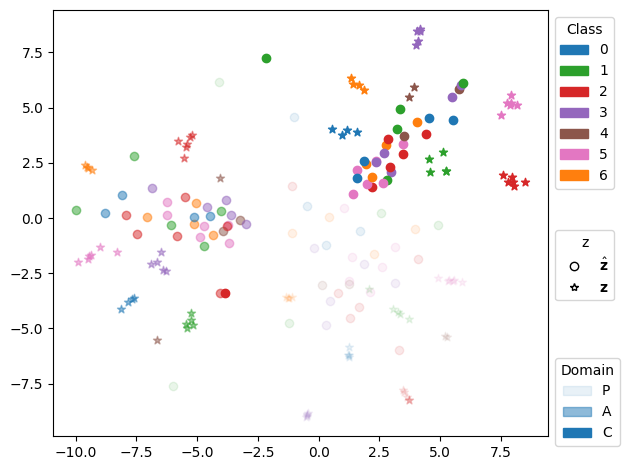}
\subcaption{$\mathbf{\hat{z}}$ without $\mathcal{L}_{\text{clsz}}$ [Eq.(\ref{eq:33})]} 
\end{minipage}
\begin{minipage}{0.24\linewidth}
\centering
\includegraphics[width=\linewidth]{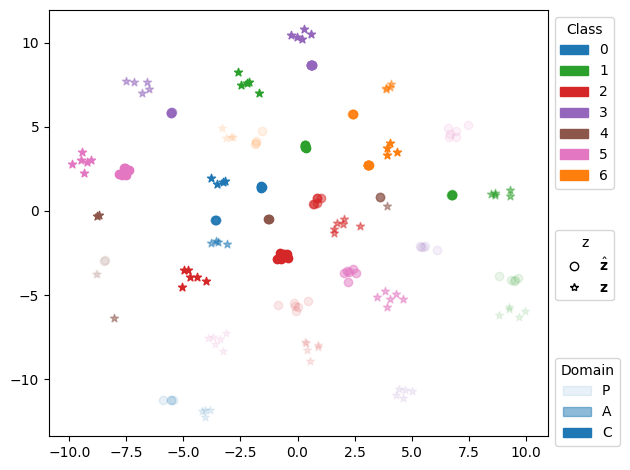}
\subcaption{$\mathbf{\hat{z}}$ without $\mathcal{L}_{\text{bn}}$ [Eq.(\ref{eq:34})]}   
\end{minipage}
\begin{minipage}{0.24\linewidth}
\includegraphics[width=\linewidth]{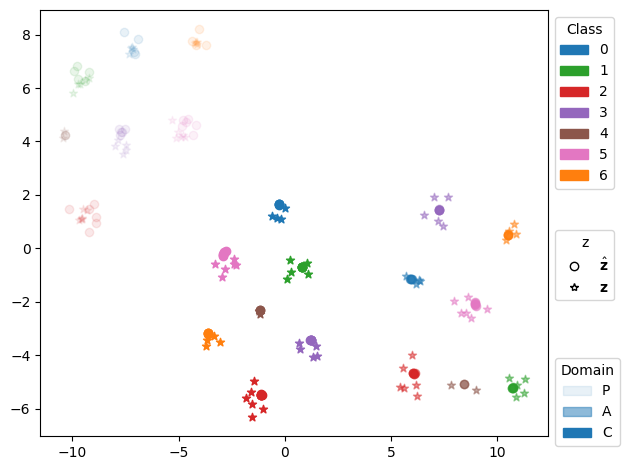}
\subcaption{$\mathbf{\hat{z}}$ without $\mathcal{L}_{\text{norm}}$ [Eq.(\ref{eq:35})]} 
\end{minipage}
\begin{minipage}{0.24\linewidth}
\centering
\includegraphics[width=\linewidth]{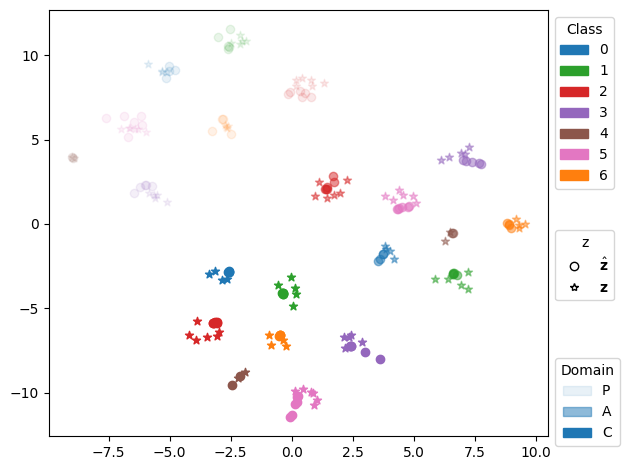}
\subcaption{$\mathbf{\hat{z}}$ with all in $\mathcal{L}_{\text{synth}}$ [Eq.(\ref{eq:36})]}
\end{minipage}
\caption{$2$ dimensional plots of original $\mathbf{z}$ ($*$ marker) and synthesized $\mathbf{\hat{z}}$ ($\circ$ marker) from PACS dataset (unseen client: \emph{sketch}, local clients: \emph{photo}, \emph{art painting} and \emph{cartoon}). Colors in the plot represent different class labels. Transparency denotes different local clients. (a) $\mathbf{\hat{z}}$ are not properly class-separated without the classification loss [Eq.(\ref{eq:33})]. (b) $\mathbf{z}$ and $\mathbf{\hat{z}}$ representations are farther in the latent space without the feature matching loss [Eq.(\ref{eq:34})]. (c)  $\mathbf{\hat{z}}$ are concentrated to few modes $\mathbf{z}$ without squared $\ell_2$ norm loss [Eq.(\ref{eq:35})]. (d) When all losses are present, $\mathbf{\hat{z}}$ are class-separated, diverse, while still remaining close to $\mathbf{z}$ [Eq.(\ref{eq:36})].}   
\label{fig:32}
\end{figure*}
Parameters that contribute more to each local client will have a greater impact on the aggregation and remain relatively unchanged when the local parameters are replaced by the global parameters in the next iteration. \emph{The stages $1$, $2$, and $3$ are only executed once. The stages $4$ and $5$ are executed iteratively,} in a similar fashion to FedAvg~\citep{mcmahan2017communication}. 
%If a parameter highly impacts multiple client models but takes a different value in each model, its weighted average could be far from its original values. If such a case arises, we average the models based on $\omega_{m}$ values as proposed. However, at the beginning of the next iteration, instead of initializing the local model parameters directly with the global model parameters, we initialize the local model with non-important parameters from the global model and retain the important parameter values from the previous iteration.
%The global model parameters initialize all local client models. 

\section{Experiments}
This section presents our experiments evaluating the proposed method. We first describe the datasets, baselines, and implementation details. Next, we report performance comparisons, ablation studies, and sensitivity analysis of the hyperparameter $\lambda_{\text{di}}$. Finally, we discuss the overhead introduced by the auxiliary components. %\setlength{\tabcolsep}{2pt}
\begin{table*}[!ht]
\centering
\caption{Unseen client accuracy on PACS and OfficeHome Datasets.}
\label{tab:pacs}
\begin{tabular}{ccccccccccc}
\toprule
Method & \multicolumn{5}{c}{PACS} &  \multicolumn{5}{c}{OfficeHome}  \tabularnewline
\cmidrule(lr){1-1}
\cmidrule(lr){2-6}
\cmidrule(lr){7-11}
& A & C & P & S & Avg &  P & A & C & R & Avg \tabularnewline
\cmidrule(lr){2-6}
\cmidrule(lr){7-11}
%\textbf{Ours} & 1 & 78.76 & 76.15& 91.14 & 73.96 & 80.00 \tabularnewline
%\textbf{Ours} & 5  & 87.70 & 84.00 & 96.41 & 78.82 & 86.73 \tabularnewline
%FedAvg~\citep{mcmahan2017communication} & 1 & 71.94 & 72.44 & 89.10 & 69.13 & 75.65\tabularnewline
% ~\citep{mcmahan2017communication}
% ~\cite{10.1007/978-3-030-58536-5_8}
% ~\citep{liu2021feddg} 
% ~\citep{chen2023federated}
% ~\citep{Zhang_2023_CVPR}
%~\citep{le2024efficiently}
FedAvg & 82.23 & 78.20 & 95.21 & 73.56 & 82.30 & 76.53 & 65.97 & 55.40 & 78.01 & 68.98  \tabularnewline
RSC & 95.21 & 83.15 & 78.24 & 74.62 & 82.81 & 75.26 & 62.34 & 50.79 & 77.46 & 66.46 \tabularnewline
ELCFS & 96.23 & 83.94 & 79.27 & 73.30 & 83.19 & 76.83 & 66.32 & 55.63 & 78.12 & 69.23 \tabularnewline
CCST &  88.33 &  78.20 & 96.65 & 82.90 & 86.52 &  76.61 &  66.35 & 52.39 & 78.01 & 68.84  \tabularnewline
FedDG-GA  & 86.91 & 81.23 & 96.80 & 82.74 & 86.92 & \textbf{77.23} & 65.10 & \textbf{58.29} & 78.80 & 69.86 \tabularnewline 
gPerXAN & 86.52 & 84.68 & 97.27 & 83.28 & 87.94 & 78.91 & 67.24 &  57.75 & 80.15 & 71.01  \tabularnewline
gPerXAN* & 74.80  & 76.47 &  87.24 & \textbf{84.34} & 80.72  & 65.20 & 52.46 & 47.12 & 65.37 & 57.54  \tabularnewline 
Ours & 88.18$\pm$0.3 & \textbf{85.12}$\pm$0.2 & \textbf{97.92}$\pm$0.2 & 81.62$\pm$0.3 & \textbf{88.21}$\pm$0.2 & 77.04$\pm$0.3  & \textbf{66.71}$\pm$0.1 & 58.23$\pm$0.2 & \textbf{79.30}$\pm$0.3 &  \textbf{70.32}$\pm$0.2 \tabularnewline

\bottomrule
\multicolumn{11}{c}{* denotes results obtained from our runs using the code provided by the author.} \tabularnewline
\end{tabular}
\end{table*}   

\begin{table*}[!ht]
\centering
\caption{Unseen client accuracy on DomainNet Dataset.}
\label{tab:dom}
\begin{tabular}{ccccccccc}
\toprule
Method & \multicolumn{6}{c}{DomainNet} \tabularnewline
\cmidrule(lr){1-1}
\cmidrule(lr){2-8}
& C & I & P & Q & R & S & Avg \tabularnewline
 \cmidrule{2-8}
%\textbf{Ours} & 1 & 78.76 & 76.15& 91.14 & 73.96 & 80.00 \tabularnewline
%\textbf{Ours} & 5  & 87.70 & 84.00 & 96.41 & 78.82 & 86.73 \tabularnewline
%FedAvg~\citep{mcmahan2017communication} & 1 & 71.94 & 72.44 & 89.10 & 69.13 & 75.65\tabularnewline
FedAvg  & 67.92 & 32.77 & 60.27 & 52.90 & 68.72 & 61.15 & 57.29  \tabularnewline
RSC &  70.96 & 34.25 & 60.31 & 55.20 & 66.91 & 63.84 & 58.58 \tabularnewline
ELCFS & 71.91 & 32.54 & 63.70 & 56.87 & 67.80 & 69.42 & 60.37 \tabularnewline
FedDG-GA & 
71.86 & \textbf{34.40} & 63.25 & 57.50 & 67.26 & 67.15 & 60.24 \tabularnewline
gPerXAN* & \textbf{75.87} & 33.92 & 67.51 & 58.74 & 69.79 & 76.84 & 63.78 \tabularnewline
\textbf{Ours} & 74.41 $\pm$ 0.3 & 31.61 $\pm$ 0.3 & \textbf{68.66} $\pm$ 0.2 & \textbf{62.93} $\pm$ 0.5 & \textbf{70.40} $\pm$ 0.2 & \textbf{77.40} $\pm$ 0.2 & \textbf{64.24} $\pm$ 0.3 \tabularnewline
\bottomrule
\multicolumn{8}{c}{* denotes results obtained from our runs using the code provided by the author.} 
\end{tabular}
\end{table*}
\subsection{Experimental Setup}
\subsubsection{\textbf{Datasets}}\label{sec5:datasets}
We evaluated our proposed method on three widely used
DG benchmarks, all of which have large discrepancies in their image styles across their domains. PACS~\citep{Li2017DeeperBA} dataset contains a total of $9,991$ images spread across four domains (\emph{art painting, cartoon, photo, sketch}).  OfficeHome~\citep{venkateswara2017deep} contains $15,588$ images split across four domains (\emph{product, art, clipart, real-world}).  DomainNet~\citep{Peng_2019_ICCV} contains $569,010$ images in six domains (\emph{clipart, infograph, painting, quickdraw, real, sketch}). 
\subsubsection{\textbf{Baselines}}\label{sec5:baselines} First, we included FedAvg~\citep{mcmahan2017communication}, a fundamental baseline under the FedDG paradigm. Then, we chose a regularization-based (centralized) DG method called RSC~\citep{10.1007/978-3-030-58536-5_8} that can be migrated to FedDG. Next, we selected ELCFS~\cite{liu2021feddg} and CCST~\cite{chen2023federated}, as these methods are related to ours because they achieve domain invariance between clients by minimizing feature divergences among clients. However, they directly share data statistics between clients to improve generalization. Finally, we also chose FedDG-GA~\citep{Zhang_2023_CVPR}, and gPerXAN~\cite{le2024efficiently}, as they are the most recent state-of-the-art methods in FedDG.
\subsubsection{\textbf{Experimental Settings}}\label{sec5:expsetting}
\paragraph{\underline{Evaluation Protocol}} We evaluated our method based on the leave-one-domain-out protocol following~\citep{chen2023federated,le2024efficiently,Zhang_2023_CVPR}, where one of the clients is chosen as the unseen client and kept aside for evaluation while the model is trained on the rest of the clients. We repeated this procedure for every domain in the dataset. We repeated the experiments three times with different seeds and reported the mean accuracy. The split of the train and the validation set within each client is kept the same as that of~\citep{gulrajani2021in} for PACS, \citep{10.1007/978-3-030-58536-5_8} for OfficeHome, and ~\citep{li2021fedbn} for DomainNet. The whole unseen client is used for testing. We selected the best model based on its performance in the validation set and reported its accuracy on the unseen client.
\setlength{\tabcolsep}{3pt}
\begin{table*}[!ht]
\centering
\caption{Global model performance on local client datasets under different aggregation strategies}
\label{tab:weight}
%\begin{adjustbox}{width=0.75\textwidth}
\begin{tabular}{cccccccccccccccc}
\toprule
Dataset & Imp. Weight Agg.? & \multicolumn{12}{c}{Performance on Local Client Data}  \tabularnewline
\midrule
%\textbf{Ours} & 1 & 78.76 & 76.15& 91.14 & 73.96 & 80.00 \tabularnewline
%\textbf{Ours} & 5  & 87.70 & 84.00 & 96.41 & 78.82 & 86.73 \tabularnewline
%FedAvg~\citep{mcmahan2017communication} & 1 & 71.94 & 72.44 & 89.10 & 69.13 & 75.65\tabularnewline
\multirow{5}{*}{PACS} &  & \multicolumn{3}{c}{Global client A} & \multicolumn{3}{c}{Global client C} &\multicolumn{3}{c}{Global client P} & \multicolumn{3}{c}{Global client S} \tabularnewline
 \cmidrule(lr){3-5}
 \cmidrule(lr){6-8}
 \cmidrule(lr){9-11}
  \cmidrule(lr){12-14}
 %& & \multicolumn{3}{c}{Local clients} & \multicolumn{3}{c}{Local clients} & \multicolumn{3}{c}{Local clients} & \multicolumn{3}{c}{Local clients} \tabularnewline
  %\cmidrule(lr){3-5}
 %\cmidrule(lr){6-8}
 %\cmidrule(lr){9-11}
  %\cmidrule(lr){12-14}
& & P & C & S & P & A & S & A & C & S  & P & A & C \tabularnewline
 \cmidrule(lr){3-5}
 \cmidrule(lr){6-8}
 \cmidrule(lr){9-11}
  \cmidrule(lr){12-14}
& No & \textbf{99.58} &96.73 &  \multicolumn{1}{c}{\textbf{96.65}} &  99.69 & \textbf{96.35} & \multicolumn{1}{c}{\textbf{96.26}} & \textbf{96.78} & 95.61 & \multicolumn{1}{c}{96.57} &  99.79 & 96.00 & 96.20 \tabularnewline  
& Yes & 99.06 & \textbf{ 96.73} & \multicolumn{1}{c}{96.44} &  \textbf{99.79} & 96.09 & \multicolumn{1}{c}{95.96} &  96.52 & \textbf{96.35} &  \multicolumn{1}{c}{\textbf{96.57}} &  \textbf{99.79} & \textbf{96.86} & \textbf{96.56} \tabularnewline
\midrule
\multirow{5}{*}{OfficeHome}& & \multicolumn{3}{c}{Global client A} & \multicolumn{3}{c}{Global client C} &\multicolumn{3}{c}{Global client P} & \multicolumn{3}{c}{Global client R} \tabularnewline
 \cmidrule(lr){3-5}
 \cmidrule(lr){6-8}
 \cmidrule(lr){9-11}
  \cmidrule(lr){12-14}
% & & \multicolumn{3}{c}{Local clients} & \multicolumn{3}{c}{Local clients} & \multicolumn{3}{c}{Local clients} & \multicolumn{3}{c}{Local clients} \tabularnewline
 %  \cmidrule(lr){3-5}
 %\cmidrule(lr){6-8}
 %\cmidrule(lr){9-11}
 % \cmidrule(lr){12-14}
&  & C & P & R  & A & P & R & A & C & R &  A & C & P \tabularnewline
 \cmidrule(lr){3-5}
 \cmidrule(lr){6-8}
 \cmidrule(lr){9-11}
  \cmidrule(lr){12-14}
& No & 82.49 & 90.38 & \multicolumn{1}{c}{\textbf{83.69}} & 75.44 & 90.54 & \multicolumn{1}{c}{84.57} & 75.67 & 81.59 & \multicolumn{1}{c}{\textbf{83.93}} & 73.58 & \textbf{82.33} & 89.18 \tabularnewline  
& Yes & \textbf{82.61} & \textbf{90.70} & \multicolumn{1}{c}{82.89} & \textbf{77.31} & \textbf{91.54} & \multicolumn{1}{c}{\textbf{84.93}} & \textbf{76.71} & \textbf{83.13} & \multicolumn{1}{c}{82.97} & \textbf{76.11} & 81.93 & \textbf{89.70} \tabularnewline
\bottomrule
\end{tabular}
%\end{adjustbox}
\end{table*}
\setlength{\tabcolsep}{2pt}
\begin{table*}
\centering
\caption{Ablation Study: Analyzing Gains from Latent Inversion vs Important Weight Aggregation}
\label{tab:abl}
%\begin{adjustbox}{width=0.9\textwidth}
\begin{tabular}{ccccccccccccc}
\toprule
\multicolumn{2}{c}{Method} & \multicolumn{5}{c}{PACS} & \multicolumn{5}{c}{OfficeHome} \tabularnewline
\cmidrule(lr){1-2}
\cmidrule(lr){3-7}
\cmidrule(lr){8-12}
$\mathcal{L}_{\text{di}}$? & Imp. Weight? & A & C & P & S & Average  & P & A & C & R & Average\tabularnewline
\cmidrule(lr){1-1}
\cmidrule(lr){2-2}
\cmidrule(lr){3-7}
\cmidrule(lr){8-12}
%\textbf{Ours} & 1 & 78.76 & 76.15& 91.14 & 73.96 & 80.00 \tabularnewline
%\textbf{Ours} & 5  & 87.70 & 84.00 & 96.41 & 78.82 & 86.73 \tabularnewline
%FedAvg~\citep{mcmahan2017communication} & 1 & 71.94 & 72.44 & 89.10 & 69.13 & 75.65\tabularnewline
No & No  & 87.55 & 83.41 & \textbf{98.06} & 78.22 & 86.81 &  \textbf{77.04} & 66.22 & 57.26 & 78.11   & 69.66 \tabularnewline
Yes & No & 88.44 & 84.87 & 97.84 & 80.29 & 87.86 & 76.75 & 65.90 & 58.12  & 78.50 & 69.82   \tabularnewline
No & Yes  & \textbf{88.47} & 84.19 & 97.56 & 77.81 & 87.01 & 76.83 & 65.78 & 57.65 & 78.96   & 69.81 \tabularnewline
Yes & Yes & 88.18 & \textbf{85.12} & 97.92 & \textbf{81.62} & \textbf{88.21} & 77.04  &  \textbf{66.71} & \textbf{58.23} & \textbf{79.30} &  \textbf{70.32} \tabularnewline
\bottomrule
\end{tabular}
%\end{adjustbox}
\end{table*}
\paragraph{\underline{\smash{Implementation Details}}}
\begin{itemize}
\item \textbf{Optimization of Local Clients (Stages $1$ and $4$)}: We modeled each encoder ($g_{\phi_d}$) as a ResNet50~\cite{he2015deepresiduallearningimage} (pretrained on Imagenet~\cite{5206848}) for PACS and OfficeHome following gPerXAN~\citep{le2024efficiently}, and AlexNet~\citep{NIPS2012_c399862d} for DomainNet following FedDG-GA~\cite{Zhang_2023_CVPR}. We modeled each classifier ($h_\theta$) as a single layer, fully connected network preceded by a $1$D batch normalization layer for all experiments. We optimized the model parameters using the SGD optimizer with a learning rate of $0.001$, momentum rate of $0.9$, weight decay of $5e^{-4}$, and a batch size of $32$. We trained each client for $10$ epochs, over $20$ communication iteration rounds. For stage $4$, we tuned $\lambda_{\text{di}}$ (hyperparameter of the domain invariance loss in Eq.(\ref{eq:314})) between [$1e^{-4}$,$10$] with a step size of $10$ and chose $\lambda_{\text{di}}=1$ from cross validation. We use the squared Euclidean norm ($\ell_2(\cdot)$ = $\parallel\cdot\parallel_2^2$) as the distance metric $\mathrm{Div}(\cdot)$ in Eq.(\ref{eq:314}). 

\item \textbf{Optimization of $\mathbf{\hat{z}}$ (Stage $2$)}: The dimension $n$ of $\mathbf{\hat{z}}$ was assigned as $512$, $512$, and $4096$ for PACS, OfficeHome, and DomainNet, respectively. We use the squared $\ell_2$ norm ($\parallel.\parallel_2^2$) as the distance metric $\mathrm{Div}(\cdot)$ in Eq.(\ref{eq:34}). We optimized $\mathbf{\hat{z}}$ using the Adam optimizer with a learning rate of $0.0001$ and a batch size of $32$ for $10,000$ epochs. We tuned $\lambda_{\text{bn}}$ (coefficient of the batch normalization loss in Eq.(\ref{eq:36})) and $\lambda_{\text{norm}}$ (coefficient of the $\ell_2$ regularization loss in Eq.(\ref{eq:36})) between [$1e^{-4}$,$1$] with a step size of $10$ and chose $\lambda_{\text{bn}}$ as $0.001$ and $\lambda_{\text{norm}}$ as $0.0001$ based on the quality of the 2D t-SNE plots of the representations in Fig~\ref{fig:32}. We synthesized $s$ samples of $\mathbf{\hat{z}}$ per source client to train the domain representation translator. Here, $s=200$ for all datasets. These samples were discarded after training the domain representation translator.

\item \textbf{Optimization of Representation Translator (Stage $3$)}: We modeled the generator and the discriminator in the representation translation network as a three-layer fully connected network with a hidden dimension of $1024$ for PACS and OfficeHome and $2048$ for DomainNet. Each layer was followed by a Leaky ReLU activation with a slope of $0.2$, a layer norm regularization, and a dropout regularization with a probability of $0.5$. We followed the optimization procedure from~\cite{NEURIPS2021_2a271795,choi2018stargan} and did not modify their hyperparameter values.
\end{itemize}
\subsection{Experimental Results}

\subsubsection{\textbf{Visualization of Synthesized Latent Representations}}\label{sec5:ts}
We visualize the original and synthesized latent representations by applying dimensionality reduction to $\mathbf{z}$ and $\mathbf{\hat{z}}$ and viewing the results in 2D using the t-SNE algorithm~\cite{van2008visualizing}. In Fig~\ref{fig:32}, we show the impact of each loss we used to synthesize $\mathbf{\hat{z}}$. From the figure, it can be seen that the synthetic representations are quite close to the originals in the latent space.
\subsubsection{\textbf{Performance on Unseen Clients}}\label{sec5:resultsm}
We report our main results on unseen client accuracy in Table~\ref{tab:pacs} for PACS and OfficeHome datasets and in Table~\ref{tab:dom} DomainNet dataset. Our method surpasses FedDG-GA, the best baseline by $1.29\%$, $0.46\%$, $4\%$ on PACS, OfficeHome, and DomainNet datasets, respectively. We had to re-run gPerXAN again because they did not provide any prior results on DomainNet dataset. Our method surpasses gPerXAN (based on our runs) by $7.49\%$, $12.78\%$, and $0.46\%$ on PACS, OfficeHome, and DomainNet\footnote{However, our runs of gPerXAN using the code provided by the author did not yield the results reported by them on PACS and OfficeHome datasets.}. On average, our method outperforms state-of-the-art FedDG-GA by $1.91\%$ and gPerXAN by $6.91\%$ (reproduced) and $0.01\%$ (reported+reproduced).
\begin{table*}[!ht]
    \centering
    \caption{Total Number of Parameters Communicated between Server and a Client $d$}
    \label{tab:commu}
    \begin{tabular}{cccccc}
    \toprule
 \multirow{3}{*}{Method} & \multirow{3}{*}{Direction}   & Models (Parameters)   & Models (Parameters)   & Total    & Total Parameters    \tabularnewline
 &  &  Communicated  &  Communicated  &   Communication &  Communicated  \tabularnewline
  &  &  [In Stages $1,2,3$] & Each Round  &   Rounds &  End of Training  \tabularnewline
\midrule
\multirow{2}{*}{FedDG-GA} & client $d$ $\rightarrow$ server & - & $\mathsf{F}_{\theta_d}$ ($\sim 23.6$M) & \multirow{2}{*}{40}  & $\sim 945$M \tabularnewline
~ &  server $\rightarrow$ client $d$ & - & $\mathsf{F}_\theta$ ($\sim 23.6$M)  &  & $\sim 945$M \tabularnewline
\midrule
\multirow{2}{*}{gPerXAN} & client $d$ $\rightarrow$ server & - & $\mathsf{F}_{\theta_d}$ ($\sim 23.7$M)  &  \multirow{2}{*}{100}  & $\sim2.37$B\tabularnewline
~ & server $\rightarrow$ client $d$ & - &  $\mathsf{F}_\theta$ ($\sim 23.7$M)  & & $\sim 2.37$B \tabularnewline
\midrule
\multirow{2}{*}{Ours} & client $d$ $\rightarrow$ server & $h_{\theta_d}$ ($\sim 33.3$K) & $g_{\phi_d}$ ($\sim 24.6$M), $h_{\theta_d}$ ($\sim 33.3$K) & \multirow{2}{*}{20} & \textbf{$\sim \textbf{492}$M} \tabularnewline
~ & server $\rightarrow$ client $d$ & $G_{\psi_G}$ ($\sim 2.1$M) & $g_\phi$ ($\sim 24.6$M), $h_\theta$ ($\sim 33.3$K)  &   & \textbf{$\sim \textbf{494}$M} \tabularnewline
\bottomrule
\end{tabular}
\end{table*}

\subsubsection{\textbf{Impact of Important Weight Aggregation on Local Data}}\label{sec5:impact} Table~\ref{tab:weight} reports the accuracy of the global model on local client data. We report the accuracy of the global model on local client data under two aggregation strategies: (i) direct averaging of local models, shown in the first row of the table, and (ii) aggregation using the important weight approach, shown in the second row. On average, aggregating local models via the important weight aggregation results in better individual local model performance than averaging without the important weight aggregation. On average, the accuracy of the global model on local client data has a gain of $0.06\%$ on PACS and $0.6\%$ on OfficeHome datasets. In the PACS dataset, $\frac{7}{12}$ times, the global model performed better on local client data with important weight averaging, while in OfficeHome, it was $\frac{9}{12}$. In particular, when the test client was the \emph{sketch} domain in PACS and the \emph{clipart} domain in OfficeHome, both among the most challenging domains to classify based on their test domain accuracies, the global model showed improved performance across all local clients when the important weight aggregation method was applied.
\subsubsection{\textbf{Ablation Study of Losses}}\label{sec5:abla}
We evaluate the impact of domain invariance loss [Eq.(\ref{eq:314})] and important weight aggregation [Eq.(\ref{eq:317})] on PACS and OfficeHome datasets in Table~\ref{tab:abl}. Without both, the average unseen client accuracy was $86.81\%$ and $69.66\%$ on PACS and OfficeHome, respectively. Using only the important weight strategy, it improved to $87.01\%$ and $69.81\%$, while domain invariance loss alone further increased it to  $87.86\%$ and $69.82\%$. This suggests domain invariance loss contributes more to generalization. However, combining both achieved the highest accuracy of $88.21\%$ and $70.32\%$. Interestingly, we observed that \emph{domain invariance loss enhances the accuracy for clients that are harder to classify (clients with the lowest accuracy in both datasets)}, indicated by the values: $(81.62\%, 80.29\%)$ over $(77.81\%, 78.22\%)$ in \emph{sketch} (PACS) and $(58.23\%, 58.12\%)$ over $(57.65\%, 57.26\%)$ in \emph{clipart} (OfficeHome). 
\subsubsection{\textbf{Analysis of the Complexity of the Auxiliary Components}} We have two auxiliary components in our algorithm compared to a general FedAvg algorithm. They are \emph{latent space inversion} and \emph{representation translation}. We analyze their computational complexity here. Let $s$ denote the total samples per client, $m$ denote the number of clients, $b$ denote the minibatch size, $n$ denote the latent space dimensionality, $\mathcal{Z}\subset\mathbb{R}^n$, and $c$ denote the number of class labels. 
For \underline{\smash{latent space inversion}}, the analysis is as follows:
\begin{itemize}
  \item \textbf{Forward pass and backward pass:} Processing $b$ samples through $h_\theta$ incurs 
        $\mathcal{O}(b\,n)$ operations. Gradients and parameter updates are computed independently for each sample, requiring $\mathcal{O}(b\,n)$ operations.%%
  \item \textbf{Losses:} Computing the \textbf{cross entropy loss} independently for $b$ outputs, yields
        $\mathcal{O}(b\,c)\approx\mathcal{O}(b)$ since $c$ is constant. \textbf{Feature matching loss} is computed using PyTorch~\cite{NEURIPS2019_9015} forward hooks. It collects \texttt{BatchNorm1d} statistics of $\mathbf{\hat{z}}$ and computes the divergence loss, costing $\mathcal{O}(b\,n)$. Evaluating a single $\lVert \mathbf{\hat{z}}\rVert_2$ (\textbf{$\ell_2$ regularization}) costs $\mathcal{O}(n)$. For $b$ samples, it would be $\mathcal{O}(b\,n)$. 
\end{itemize}
Even though aggregating these operations over all samples ($s$), clients ($m$), and gradient iterations ($g$) increases the run time, the asymptotic total computational complexity scales linearly with respect to the input size $b$ as they can be operated on in parallel. 
For the \underline{\smash{representation translator}}, designed with $\text{l}$ fully connected layers and hidden width of $w$, the complexity is as follows.
\begin{itemize}
    \item \textbf{Forward pass and backward pass for $G$ and $D$:} If the input layer maps from $n$ to $w$, the first layer contributes $\mathcal{O}(b\,n\,w)$, and similarly, the final output layer contributes $\mathcal{O}(b\,w\,n)$ (for $G$) and $\mathcal{O}(b\,w)$ (for $D$), and the intermediate layers contribute $\mathcal{O}(b\,\text{l}\,w^2)$. Thus, the total cost is $\approx \mathcal{O}(b\,\text{l}\,w^2)$. 
    \item \textbf{Loss:} The cost of cross entropy loss is $\mathcal{O}(b\,c)\approx\mathcal{O}(b)$. The adversarial and reconstruction losses contribute a complexity of $\mathcal{O}(b\,\text{l}\,w^2)$. The client classification loss adds a cost of $\mathcal{O}(b\,w\,m)$.
\end{itemize}
\begin{table}[!ht]
\centering
\caption{Parameters Stored by Our Method at Each Stage}
\label{tab:storage}
%\begin{adjustbox}{width=\textwidth}
\begin{tabular}{cccccc}
\toprule
\multicolumn{3}{c}{Stages 1, 2, and 3} & \multicolumn{2}{c}{Stages 4 and 5} & After Training \tabularnewline
\midrule
Model & Stored At & \# Parameters & Model & Stored At  & Model \tabularnewline
\midrule
$g_{\phi}$ & server & $ 24.6$M &$g_{\phi}$  & server & $g_{\phi}$   \tabularnewline
$g_{\phi_d}$ & client & $ 24.6$M &$g_{\phi_d}$  & client  & $g_{\phi_d}$  \tabularnewline
$h_{\theta_d}$, $h_{\theta_d}^{\ddag}$ & client, server & $33.3$K & $h_{\theta_d}$ & client &$h_{\theta_d}$  \tabularnewline
$h_{\theta}$ &  server & $ 33.3$K & $h_{\theta}$ & server & $h_{\theta}$  \tabularnewline
$G_{\psi_G}$ &  server & $ 2.1$M & $G_{\psi_G}^{\dag}$ & client   \tabularnewline
$D_{\psi_\text{src}}^{\S}$,$D_{\psi_\text{cls}}^{\S}$ & server &  $ 1.0$M  & ~ & ~ & ~ \tabularnewline
 $\mathbf{\hat{z}}^{\S}$ &  server  & $9.8$M & ~ & ~ & ~  \tabularnewline
\bottomrule
\multicolumn{6}{c}{$\ddag$ purged after Stage 2, $\S$ purged after Stage 3, and $\dag$ purged after Stage 5 }
    \end{tabular}
%\end{adjustbox}
\end{table}
These operations also scale linearly with respect to the input size $b$. These auxiliary modules are implemented using shallow fully connected networks operating on low-dimensional latent vectors. As a result, including these components only introduces a marginal computational overhead. For analysing \underline{\smash{space complexity}}, we report the parameters needed by our method at each stage in Table~\ref{tab:storage}. It can be seen that the latent representations and the discriminator can be purged after stage $3$. The representation translator can also be purged at the end of model training (stage $5$). Hence, no extra storage is required for our method once the model is trained. 
\begin{figure}
\centering
\captionsetup[subfigure]{justification=centering}
\begin{minipage}{0.3\linewidth}
\includegraphics[width=\linewidth]{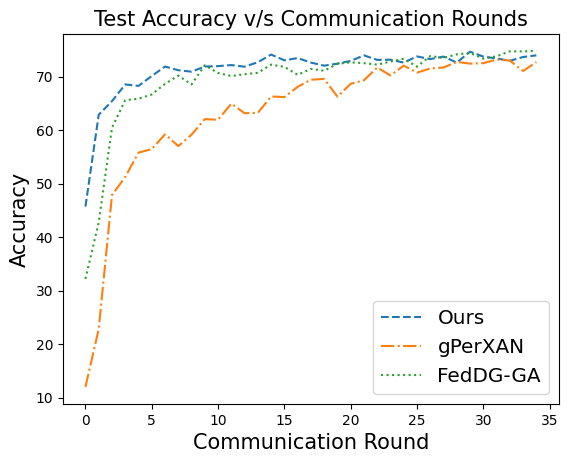}
\subcaption{client C} 
\end{minipage}
\begin{minipage}{0.3\linewidth}
\includegraphics[width=\linewidth]{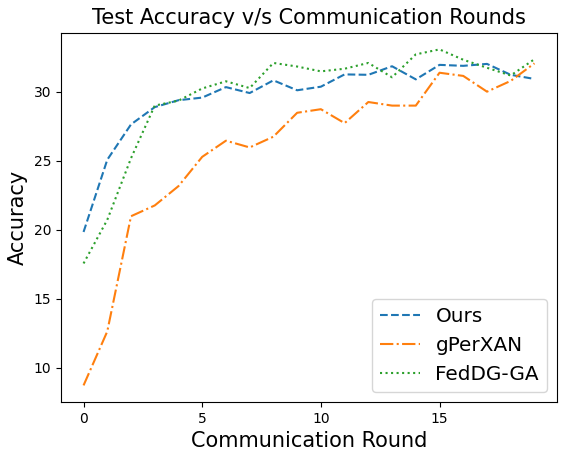}
\subcaption{client I} 
\end{minipage}
\begin{minipage}{0.3\linewidth}
\includegraphics[width=\linewidth]{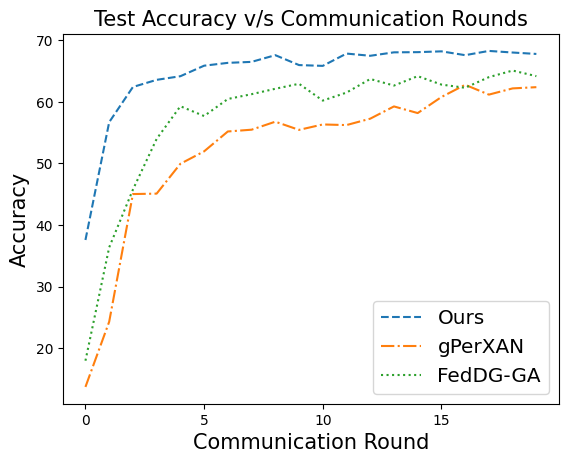}
\subcaption{client P} 
\end{minipage}
\begin{minipage}{0.3\linewidth}
\centering
\includegraphics[width=\linewidth]{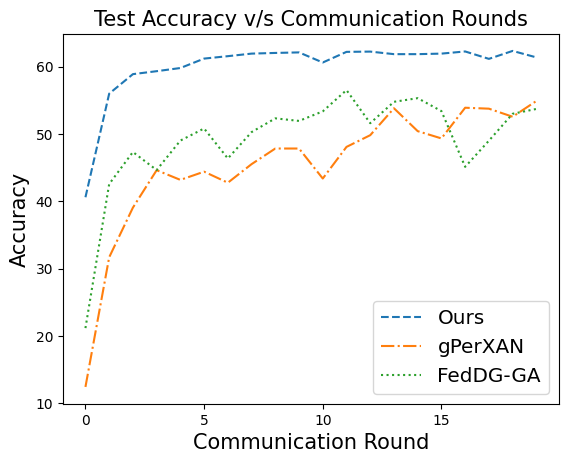}
\subcaption{client Q}  
\end{minipage}
\begin{minipage}{0.3\linewidth}
\includegraphics[width=\linewidth]{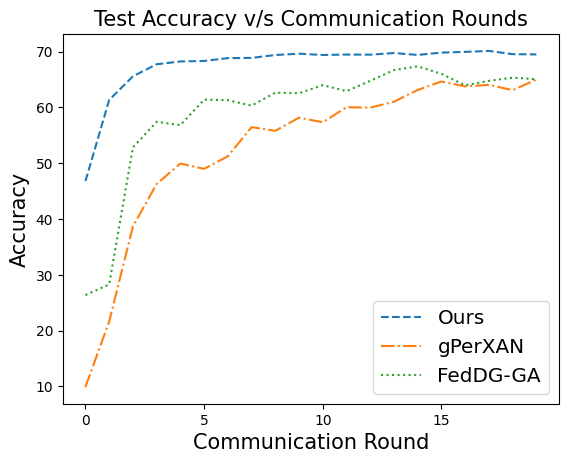}
\subcaption{client R}
\end{minipage}
\begin{minipage}{0.3\linewidth}
\centering
\includegraphics[width=\linewidth]{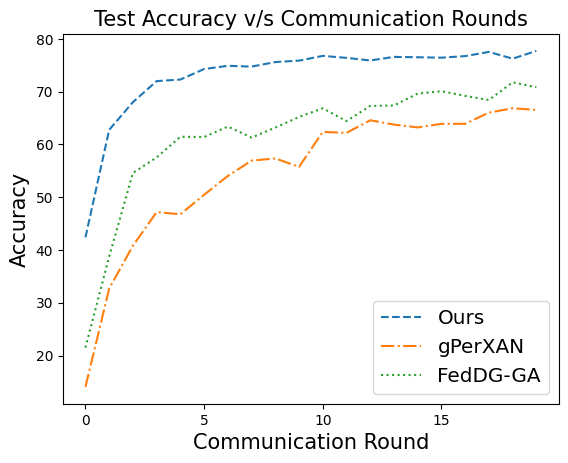}
\subcaption{client S} 
\end{minipage}
\caption{Relationship between \emph{unseen client accuracy} (Y-axis) and communication rounds (X-axis) on different clients in DomainNet. In almost all the cases, our method (in blue) empirically converges to higher accuracy in significantly fewer communication rounds than baselines gPerXAN (in orange) and FedDG-GA (in green), making it communication efficient.} 
\label{fig:33}
\end{figure}
\subsubsection{\textbf{Analysis on Communication Overhead}}\label{sec5:anal}
Fig~\ref{fig:33} shows that our method converges to the maximum unseen client accuracy within $5$ communication rounds, while it takes about $20$ rounds for the baseline methods (gPerXAN and FedDG-GA) to reach close to that level of accuracy in multiple test clients of DomainNet dataset. For instance, let us consider test client P (\emph{painting}). After a single round of communication, the test client accuracy of our global model is $37.57\%$ compared to $13.69\%$ for gPerXAN and $17.95\%$ for FedDG-GA, a difference of almost $20\%$. This trend is prevalent across all clients. After $5$ rounds, the test client accuracy of our global model is $64.18\%$ compared to $49.95\%$ for gPerXAN and $59.30\%$ for FedDGA. After $15$ rounds, our model is at $68.09\%$ while gPerXAN is at $58.19\%$, and FedDG-GA is at $64.19\%$. This result can be extrapolated to other clients, too. Therefore, our method achieves higher performance with significantly lower communication overhead than other approaches.

In Table~\ref{tab:commu}, we analyze the total number of parameters communicated between the server and a single client $d$ for the OfficeHome dataset. Although an additional communication step was introduced during stages $1-3$, our model converged faster than other methods. This allowed us to reduce the overall volume of parameter communications between the server and client.
\begin{table}[!ht]
\centering
\caption{Sensitivity analysis on hyperparameter $\lambda_{\text{di}}$ on PACS dataset}
\label{tab:lambda2}
\begin{tabular}{ccccccc}
\toprule
$\lambda_{\text{di}}$ & \multicolumn{5}{c}{PACS} \tabularnewline
\midrule
& A & C & P & S & Average \tabularnewline
\cmidrule(lr){2-6}
%\textbf{Ours} & 1 & 78.76 & 76.15& 91.14 & 73.96 & 80.00 \tabularnewline
%\textbf{Ours} & 5  & 87.70 & 84.00 & 96.41 & 78.82 & 86.73 \tabularnewline
%FedAvg~\citep{mcmahan2017communication} & 1 & 71.94 & 72.44 & 89.10 & 69.13 & 75.65\tabularnewline
10 & 87.16 & 85.09 & \textbf{98.06} & 80.57 & 87.72 \tabularnewline
1 & 88.18 & \textbf{85.12} & 97.92 & \textbf{81.62} & 88.21 \tabularnewline
0.1 & 87.85 & \textbf{85.12} & 97.76 & 81.18 & 87.98 \tabularnewline
0.01 & 87.94 & 84.15 & 97.88 & 77.90 & 86.97 \tabularnewline
0.001 & \textbf{88.33} & 84.66 & 97.96 & 78.83 & 87.45 \tabularnewline
0.0001 & 87.60 & 83.99 & 97.86 & 79.32 & 87.19 \tabularnewline
\bottomrule
\end{tabular}
\end{table}
\subsubsection{\textbf{Sensitivity Analysis on the Hyperparameter}}\label{sec5:sen}
Table~\ref{tab:lambda2} shows how global model accuracy varies based on changes in hyperparameter $\lambda_{\text{di}}$ values in Eq.(\ref{eq:315}). On average, higher $\lambda_{\text{di}}$ values contributed to better test accuracies, demonstrating the significance of domain invariance loss.

\section{Conclusion}
We introduced a novel approach to establish domain invariance in the federated domain generalization paradigm. We propose a method to synthesize data in the latent space to overcome the lack of centralized data and safeguard the potential risk of privacy. By optimizing the latent representations in each local model to be domain invariant with respect to the synthesized latent representations, we minimize the divergences between the client distributions, leading to local and global models with better generalization capability. We introduced a new strategy to aggregate local models to prevent any loss of local adaptations. We also demonstrated that our approach results in a higher accuracy with limited communication overhead compared to the state-of-the-art baselines.  

%\section{Acknowledgment}
%This research was partially supported by the Australian Government through the Australian
%Research Council’s Discovery Projects funding scheme (project DP210102798). The views expressed herein are those of the authors and are not necessarily those of the Australian Government or the Australian Research Council.
\bibliographystyle{unsrt2authabbrvpp}
\bibliography{ref}
\end{document}